\documentclass[sigconf,anonymous=False]{acmart}

\AtBeginDocument{%
  \providecommand\BibTeX{{%
    \normalfont B\kern-0.5em{\scshape i\kern-0.25em b}\kern-0.8em\TeX}}}
    
\usepackage{graphicx}
\usepackage{subcaption}
\usepackage{wrapfig}
\usepackage{dsfont}

\setcopyright{acmcopyright}
\copyrightyear{2020}
\acmYear{2020}
\acmDOI{_}

\acmConference[ICAIF '20]{ACM International Conference on AI in Finance}{October 15-16, 2020}{New York, USA}
\acmBooktitle{}
\acmPrice{15.00}
\acmISBN{}

\begin{document}

\title{Learning Sampling in Financial Statement Audits using Vector Quantised Autoencoder Neural Networks}

%%
%% The "author" command  and its associated commands are used to define
\author{Marco Schreyer}
\affiliation{%
    \institution{University of St. Gallen}
    \city{St. Gallen}
    \country{Switzerland}
}
\email{marco.schreyer@unisg.ch}

\author{Timur Sattarov}
\affiliation{%
    \institution{Deutsche Bundesbank}
    \city{Frankfurt am Main}
    \country{Germany}
}
\email{timur.sattarov@bundesbank.de}

\author{Anita Gierbl}
\affiliation{%
    \institution{University of St. Gallen}
    \city{St. Gallen}
    \country{Switzerland}
}
\email{anita.gierbl@unisg.ch}

\author{Bernd Reimer}
\affiliation{%
    \institution{PricewaterhouseCoopers GmbH WPG}
    \city{Stuttgart}
    \country{Germany}
}
\email{reimer.bernd@pwc.com}

\author{Damian Borth}
\affiliation{%
    \institution{University of St. Gallen}
    \city{St. Gallen}
    \country{Switzerland}
}
\email{damian.borth@unisg.ch}

%\affiliation{%
%  \institution{Palmer Research Laboratories}
%  \streetaddress{8600 Datapoint Drive}
%  \country{Texas}

\renewcommand{\shortauthors}{Schreyer and Sattarov, et al.}

\begin{abstract}

The audit of financial statements is designed to collect reasonable assurance that an issued statement is free from material misstatement ’true and fair presentation’. International audit standards require the assessment of a statements' underlying accounting relevant transactions referred to as 'journal entries' to detect potential misstatements. To efficiently audit the increasing quantities of such entries, auditors regularly conduct a sample-based assessment referred to as 'audit sampling'. However, the task of audit sampling is often conducted early in the overall audit process. Often at a stage, in which an auditor might be unaware of all generative factors and their dynamics that resulted in the journal entries in-scope of the audit. To overcome this challenge, we propose the application of \textit{Vector Quantised-Variational Autoencoder (VQ-VAE)} neural networks. We demonstrate, based on two real-world city payment datasets, that such artificial neural networks are capable of learning a quantised representation of accounting data. We show that the learned quantisation uncovers (i) the latent factors of variation and (ii) can be utilised as a highly representative audit sample in financial statement audits. % Initial feedback received by auditors underpinned the effectiveness of the approach. 

\end{abstract}




\maketitle

\section{Introduction}
\label{sec:introduction}

\begin{figure}[t!]
	\hspace*{0.0cm} \includegraphics[width=8.5cm, angle=0, trim={1.0cm 2.0cm 0.0 0.0}]{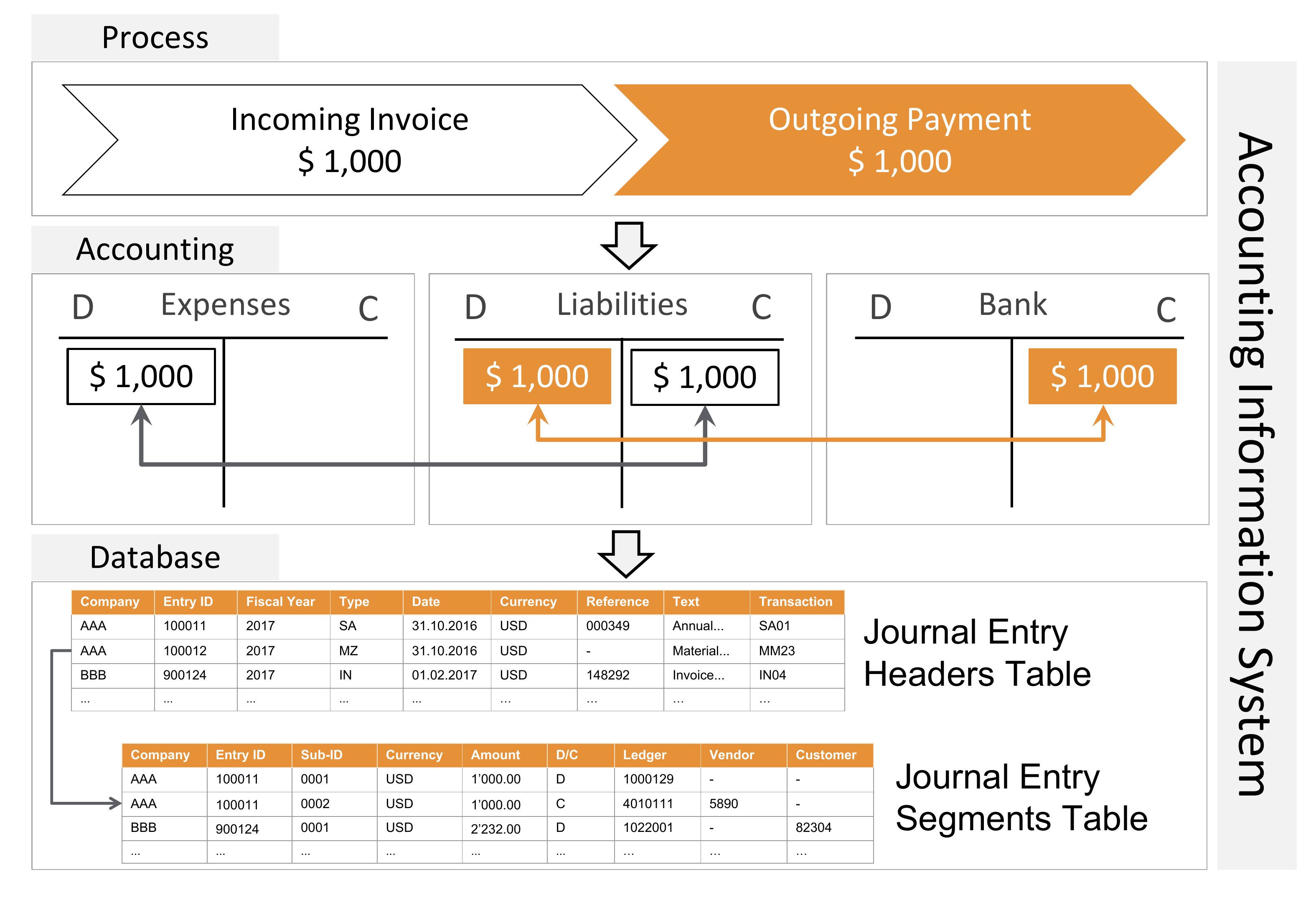}
	\caption{Hierarchical view of an Accounting Information System (AIS) that records distinct layer of abstractions, namely (1) the business process, (2) the accounting and (3) technical journal entry information in designated tables.}
	\label{fig:ais_system}
\end{figure}

The trustworthiness of financial statements plays a fundamental role \citep{ias2007} in today's economic decision making by investors. The audit of such statements, conducted by external auditors, is designed to collect reasonable assurance that an issued financial statement is free from material misstatement (’true and fair presentation’) \citep{aicpa2002}, \citep{ifac2009}. International audit standards require an assessment of the statement's underlying accounting relevant \textit{journal entries} to detect a potential misstatement \citep{caq2018}. Journal entries debit and credit the distinct accounting ledgers of a financial statement evident in its balance sheet and profit and loss statement. Nowadays, organizations collect vast quantities of such journal entries in \textit{Accounting Information Systems (AIS)} or more general \textit{Enterprise Resource Planning (ERP)} systems \citep{grabski2011}. Figure \ref{fig:ais_system} depicts a hierarchical view of the journal entry recording process in designated database tables of an ERP system.

To efficiently audit the increasing journal entry quantities, auditors regularly conduct a sample-based assessment referred to as \textit{audit sampling}. Formally, audit sampling is defined as the \textit{'selection and evaluation of less than 100\% of the entire population of journal entries'} \citep{aicpa2012}. While sampling increases the efficiency of a financial audit, it also increases its \textit{sampling risk}. The term sampling risk denotes the likelihood that the auditor's conclusion based on auditing a subset of entries may differ from the conclusion of auditing the entire population \citep{guy2002}. Auditors are required to select a \textit{representative} sample from the population of journal entries \citep{isa2009a} to mitigate sampling risks. The selection of such a representative sample is determined by the design of the applied sampling technique. International audit standards require auditors to built their audit approach on either (i) \textit{non-statistical} or (ii) \textit{statistical} sampling techniques. Non-statistical or 'judgemental' sampling denotes the selection of audit samples based on the auditor's experience, inherent risk assessment, and 'professional judgment'. Statistical sampling, in contrast, aims to provide greater sampling objectivity. It refers to the random selection of audit samples and the determination of probabilities to evaluate the sampling results \citep{isa2009a}. To conduct statistical sampling and determine a representative sample size, auditors nowadays use pre-calculated probability tables or sampling functions available in computer-assisted audit software \citep{schwartz1998}, such as ACL\footnote{https://www.wegalvanize.com},  IDEA\footnote{https://idea.caseware.com}, or proprietary applications.

% Statistical sampling encompasses (i) \textit{attribute sampling} e.g. to assess the effectiveness of a particular internal control and (ii) \textit{variable sampling} e.g. to determine an overstatement of inventories or accounts receivables \citep{guy2002}.

% This, often without being in possession of a detailed knowledge of the journal entries latent generative factors (e.g., detailed business process workflows, granular business process characteristics).  awareness potential structural breaks (e.g., control, workflow or personnel changes) that occurred in the fiscal year subject to audit.

% Nowadays, due to the increase in volume, auditors substantively test a sample size of less than one percent of all journal entries \citep{no2019}. Regulatory bodies, such as the U.S. 'Public Company Accounting Oversight Board' (PCAOB), often question if such a limited sample size can be considered representative \cite{pcaob2019}.

During an annual audit, the task of audit sampling is regularly conducted early in the audit process. Thereby, auditors need to decide on sensitive sampling parameters before conducting the sampling, e.g. the materiality levels, confidence intervals, acceptable risk thresholds \cite{jokovich2013}. Even though the auditor might be unaware of all latent factors that generated the journal entries in-scope of the audit, e.g. the underlying business processes and workflows. This challenge is in particular evident in the context of new audit engagements in which auditors are mandated to audit an organization's financial statement for the first time. However, it is also of relevance in mature audit engagements. Especially in scenarios where the generative factors of an organization vary dynamically over time \cite{reichert2012}, e.g. due to an organizational merger or carve-out and resulting corresponding process- and workflows-changes.

% that occurred within the in-scope time period of the audit. a journal entries population

Driven by the rapid technological advances of artificial intelligence in recent years, techniques based on deep learning \citep{LeCun2015} have emerged into the field of finance \citep{lopez2018}, \citep{wiese2020} and particular financial statement audits \citep{schreyer2019a}, \citep{schultz2020}, \citep{schreyer2019b}, \citep{bhattacharya2020}. These developments raise the question: Can such techniques also be utilized for learning representative audit samples? And can the samples be drawn to be interpretable by a human auditor? In this work, we propose the application of \textit{Vector Quantised-Variational Autoencoder (VQ-VAE)} neural networks \cite{vanDenOord2017} to address those questions. Inspired by the ideas of discrete neural dimensionality reduction, we will demonstrate how VQ-VAE neural networks can be trained to learn (i) representative and (ii) human interpretable audit samples. In summary, we present the following contributions:

\begin{itemize}

\item We demonstrate that VQ-VAEs can be utilized to learn a low-dimensional representation of journal entries recorded in AIS systems that disentangle the entries latent generative factors of variation;

\item We show that the VQ-VAEs training can also be regularized to learn a set of discrete embeddings that quantise the representations generative factors and therefore constitute a representative audit sample;

\item We illustrate that the learned quantization corresponds to the entries accounting process characteristics and provides a starting point for human interpretable audit sampling for downstream substantive audit procedures.

% \item We show that the learned representation combined with the magnitude of an entry's reconstruction error can be interpreted as a highly adaptive anomaly assessment of journal entries.

\begin{figure*}[ht!]
    \center
    \includegraphics[height=6.0cm]{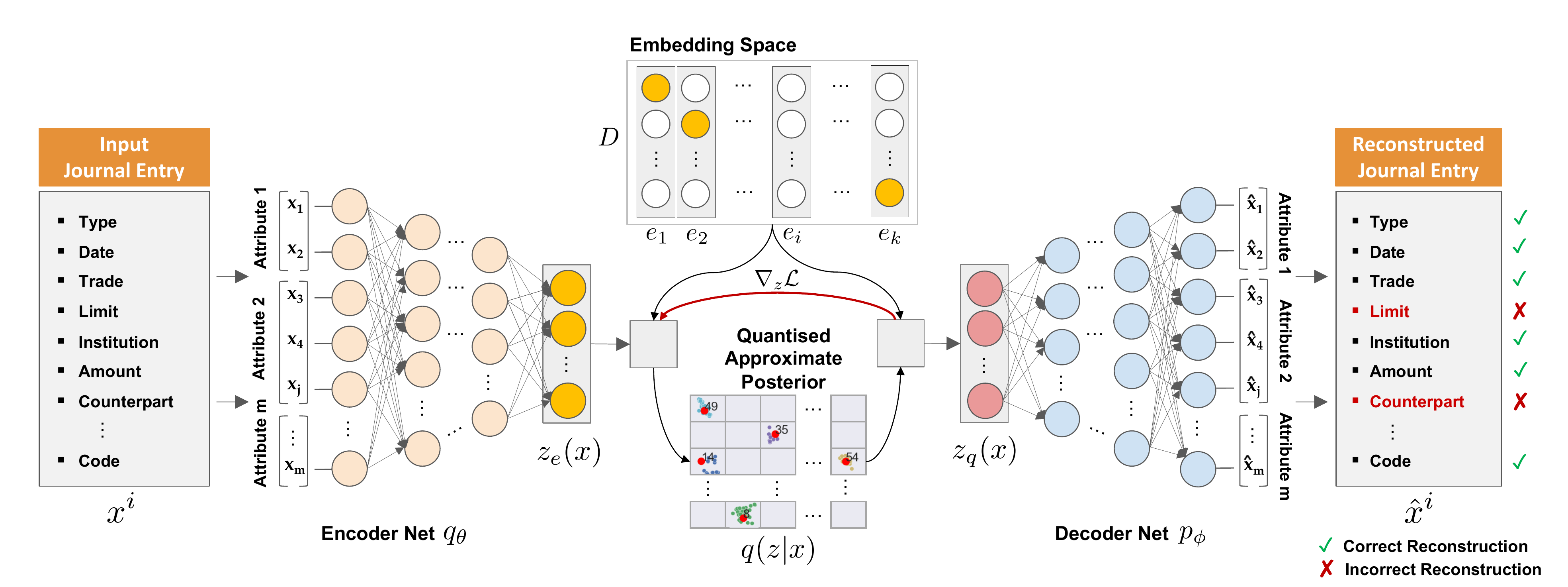}
    \caption{The VQ-VAE architecture \cite{vanDenOord2017}, applied to learn audit sampling. During the models forward pass, an input journal entry $x^{i}$ is passed through the encoder network $q_{\theta}$ producing a latent representation $z_{e}$. The VQ-VAE quantises $z_{e}$ using a \textit{codebook} of embedding vectors $e_{j}$. Afterwards, the quantised embedding $z_{q}$ is passed to the decoder network $p_{\phi}$ to reconstruct the journal entry $\hat{x}^{i}$ as faithfully as possible.}
    \label{fig:architecture}
\end{figure*}

\end{itemize}

%We hypothesize that such entries $X$ are generated by 'latent' factors of variation in $Z$ that can be recovered by an unsupervised deep learning algorithm \citep{bengio2013}.

%In the context of the rapid development of deep adversarial learning, this observation raises the following questions: Are financial audits vulnerable to adversarial attacks? And if so, are state-of-the-art CAATs able to detect such attacks? In this paper, we present a deep learning-based adversarial attack against CAATs. We regard this work to be an initial step towards the investigation of such future challenges in financial audits. In summary, we present the following contributions: First, we describe a real-world 'threat model' designed to camouflage accounting anomalies or fraudulent activities using adversarial entries. Second, we show that deep neural networks are capable of learning models of journal entries that semantically disentangle their latent generative factors. Third, we demonstrate how a potential perpetrator can maliciously misuse such learned models to create 'adversarial' journal entries. The created entries are deliberately designed to misguide the auditors CAATs during a financial audit.

% We envision this deep learning-based sampling methodology as an important supplement to toolbox of auditors \cite{Pedrosa2014}. 

The remainder of this work is structured as follows: In Section \ref{sec:relatedwork}, we provide an overview of related work. Section \ref{sec:methodology} follows with a description of the VQ-VAE architecture and presents the proposed methodology to learn a representative sample from vast quantities of journal entries. The experimental setup and results are outlined in Section \ref{sec:experiments} and Section \ref{sec:results}. In Section \ref{sec:summary}, the paper concludes with a summary of the current work and future research directions. % A reference implementation of the proposed methodology will be made available via [url redacted due to double blind review]. % \url{https://github.com/GitiHubi/deepAQ}.

\section{Related Work}
\label{sec:relatedwork}

Due to its high relevance for the financial audit practice, the task of audit sampling triggered a sizable body of research by academia \citep{akresh1988}, \citep{elder2013} and practitioners \citep{guy1998}, \citep{hitzig1995}. In the realm of this work, we focus our literature review on the two main classes of statistical sampling techniques, namely (i) attribute sampling, and (ii) variables sampling used nowadays \cite{hall2002} in auditing journal entries:

\subsection{Attribute Sampling Techniques}
\label{sec:attribute_sampling}

The technique of attribute sampling is applied by auditors to estimate the percentage of journal entries that possess a specific attribute or characteristic. The different techniques of attribute sampling encompass the following classes \citep{guy2002}:  

\begin{itemize}

\item \textit{Random sampling} in which each journal entry of the population has an equal chance of being selected.

\item \textit{Systematic} or \textit{sequential sampling} in which the sampling starts by selecting an entry at random and then every $n$-th journal entry of an ordered sampling frame is selected.

\item \textit{Proportional, block} or \textit{stratified sampling} in which the population is sub-divided into homogeneous groups of journal entries to be sampled from \citep{liu2005}.

\item \textit{Haphazard sampling}, in which no explicit or structured selection strategy is employed by the auditor. The sampling is conducted without a specific reason for including or excluding journal entries \citep{hall2013}.

\end{itemize}

In general, attribute sampling is utilised by auditors to test the effectiveness of internal controls, e.g. the percentage of vendor invoices that are approved in compliance with the organisation's internal controls, e.g. by following a 'four-eye' approval principle.

\subsection{Variable Sampling Techniques}
\label{sec:variable_sampling}

The technique of variable sampling is applied by auditors to estimate the amount or value of specific journal entry characteristics. The different techniques of variable sampling encompass the following classes \citep{guy2002}: 

\begin{itemize}

\item \textit{Difference estimation} calculates the average difference between audited amounts and recorded amounts to estimate the total audited amount of a population \citep{roberts1978}.

\item \textit{Ratio estimation} calculates the ratio of audited amounts to recorded amounts to estimate the total dollar amount of the journal entry population \citep{garstka1979}. 

\item \textit{Mean-per-unit estimation} projects the sample average to the total population by multiplying the sample average by the number of items in a journal entry population \citep{neter1975}.

\item \textit{Monetary-} or \textit{Dollar-unit sampling} considers each monetary unit as a sampling unit of the journal entry population. As a result, entries that record-high posting amounts exhibit a proportionally higher likelihood of being selected \citep{stringer1963}, \citep{leslie1979}.

\end{itemize}

In general, variable sampling is utilized by auditors in the context of substantive testing procedures, e.g. the overstatement of accounts receivables. 

To the best of our knowledge, this work presents the first deep-learning inspired approach to learn representative and human interpretable audit samples from real-world accounting data.

% Several references propose different audit sampling approaches that reduce sampling risk and thereby seek to improve the efficiency of financial statement audits.

\section{Methodology}
\label{sec:methodology}

In this section, we describe the architectural setup of the VQ-VAE model \citep{vanDenOord2017} used to learn representative audit sampling. Furthermore, we provide details on the objective function applied to optimize the model parameters.

\subsection{Latent Generative Factors}
\label{sec:entries}

Let $X$ formally be a set of $N$ journal entries $x^{1}, x^{2}, ..., x^{n}$, where each journal entry $x^{i}$ consists of $M$ accounting specific attributes $x_{1}^{i}, x_{2}^{i}, ..., x_{j}^{i}, ..., x_{m}^{i}$. The individual attributes $x_{j}$ describe the journal entries details, e.g., the entries' fiscal year, posting type, posting date, amount, general-ledger. Following the theoretical assumptions of unsupervised representation learning \citep{bengio2013a} we assume that each entry $x_{j}^{i}$ is generated by distinct factors of variation. The different variational factors are not directly observable and correspond to manifolds in a latent space $Z$. We hypothesise that each generative latent factor $z_{i} \in Z$ corresponds to a behavioural posting pattern. As a result, it can be uncovered by an unsupervised deep learning algorithm \citep{radford2015}, \citep{chen2016}.

\subsection{VQ-VAE Model}
\label{sec:vqvaemodel}

It is often intractable to directly calculate the exact posterior distribution $q(z)$ over the latent generative factors in $Z$. In \citep{kingma2013} Kingma and Welling proposed the Variational Autoencoder (VAE) to learn an approximation of the intractable posterior distribution $q(z|x)$ given the input data $X$. The VAE's encoder network $q_{\theta}$ learns to parameterise a continuous approximate posterior $q(z|x)$ over the latent factors $Z$. In parallel, the VAE's decoder network $p_{\phi}$ learns to reconstruct the input data $X$ as faithfully as possible with distribution $p(\hat{x}|z)$ over the reconstructions $\hat{X}$. To quantise the approximate posterior $q(z|x)$ and thereby learn a representative audit sample we apply the VQ-VAE model introduced by Van den Oord et al. in \citep{vanDenOord2017} and as shown in Fig. \ref{fig:architecture}. In contrast to the VAE, the VQ-VAE model defines a discrete latent embedding space $E \in \mathcal{R}^{KxD}$ where $K$ denotes the size of the space, and $D$ is the dimensionality of each discrete latent embedding vector $e_{j} \in E$. Due to its discrete nature the embedding space encompasses in total $K$ distinct embedding vectors $e_{j} \in \mathcal{R}^{D}, j= 1, 2, ..., K$. During the models forward pass an input journal entry $x^{i}$ is passed through the encoder network $q_{\theta}$ producing a latent representation $z_{e}(x)$. To obtain its corresponding quantised representation $z_{q}(x)$, the nearest neighbor lookup is performed, as defined by:

\begin{equation}
    z_{q}(x) = e_{k}, \text{  where  } k = \arg \min_{j} ||z_{e}(x) - e_{j}||_{2}.
\label{equ:embedding}
\end{equation} 

\noindent where $z_{e}(x)$ denotes the output of the encoder network and $e_{j}$ the distinct embedding vectors. This process can be viewed as passing each $z_{e}(x)$ through a discretisation bottleneck by mapping it onto its nearest embedding in $E$. Thereby the VQ-VAE autoencoder quantises the representations $z_{e}(x)$ using a \textit{codebook} of 1-of-K embedding vectors $e_{j}$. Afterwards, the quantised embedding $z_{q}(x)$ is passed to the decoder. The complete set of parameters of the model are the union of the parameters of the encoder, decoder, and the embedding space. The learned quantised posterior distribution $q(z|x)$ probabilities are defined as 'one-hot', given by: 

\begin{equation}
q(z=k|x) = \begin{cases}
            1 & \text{ for  } k = \arg \min_{j} ||z_{e}(x)-e_{j}||_{2}, \\
            0 & \text{ otherwise, }
            \end{cases}
\label{equ:lookup}
\end{equation} 

\noindent where $z_{e}(x)$ denotes the output of the encoder network and $e_{j}$ denotes a particular embedding vector.

\subsection{VQ-VAE Learning}
\label{sec:vqvaetraining}

In the forward pass, to learn a set of quantised embeddings of real-world journal entry data, we compute the model loss $\mathcal{L}$ as defined in Eq. \ref{equ:trainloss}. The loss function is comprised of four terms that are optimised in parallel to train the different model components of the VQ-VAE, given by:

\begin{equation}
\begin{aligned}
    \mathcal{L}(x) = {} & \log p(x|z_{q}(x))\;+\;\alpha ||sg[z_{e}(x)] - e||_{2}^{2} \\
    &+\;\beta ||z_{e}(x) - sg[e]||_{2}^{2}\;+\;\gamma \log p(x|z_{e}(x)), 
\end{aligned}
\label{equ:trainloss}
\end{equation} 

\noindent where $z_{e}(x)$ denotes encoder output, $z_{q}(x)$ the quantised encoder output and $e$ the set of embedding vectors. The first term denotes the discrete reconstruction loss derived from the reconstruction of the quantised embeddings $z_{q}(x)$. It encourages the embeddings $z_{q}(x)$ to be an informative representation of the input \citep{roy2018}. Throughout the training process, the embeddings $e$ receive no gradients from the reconstruction loss. This originates from the straight-through gradient estimation $\nabla_{z} \mathcal{L}$ of mapping $z_{e}(x)$ to its nearest neighbor $z_{q}(x)$. The embeddings are optimized using a vector quantization technique that applies a \textit{stop-gradient} operator denoted by $sg[\cdot]$. The operator is defined as the identity in the forward pass and has zero partial derivatives in the backward pass \citep{vanDenOord2017}, denoted by: 

\begin{equation}
sg[x] = \begin{cases}
            x & \text{ forward pass }, \\
            0 & \text{ backward pass },
            \end{cases}
\label{equ:stopgradient}
\end{equation} 

\noindent where $x$ denotes the input vector to the stop-gradient operator. 

Both the second and third loss term use stop-gradients. The second term denotes the \textit{embedding loss}, that moves the embeddings $e$ towards the encoder outputs $z_{e}(x)$. Due to the non-differentiability of the embedding assignment, the embedding space is dimensionless. It can grow arbitrarily if the embeddings $e$ do not train as fast as the encoder parameters ${\theta}$. Therefore, the third term of Eq. \ref{equ:trainloss} denotes the \textit{commitment loss} which guarantees that the encoder output $z_{e}(x)$ commits to one of the embeddings. To encourage the encoder output $z_{e}(x)$ to remain an informative representation and not 'collapsing' towards one of the embeddings we enhanced $\mathcal{L}$ by a fourth loss term as recently introduced by Fortuin et al. in \cite{fortuin2018}. The added term, denotes the reconstruction loss derived from the encoder outputs $z_{e}(x)$. The gradient $\nabla_{z} \mathcal{L}$ of the backward pass is approximated similar to the straight-through estimator presented in \citep{bengio2013b}. Thereby, the gradients of the decoder input $z_{q}(x)$ are copied back to the encoder output $z_{e}(x)$ as shown in Fig. \ref{fig:architecture}. Since the output representation of the encoder and the input to decoder share the same $D$ dimensional space, the gradients contain useful information on how to update the encoder parameters $\theta$ to increase the model likelihood. The VQ-VAE can be interpreted a special case of the VAE in which the model's likelihood $\log p(x)$ can be maximised by the optimisation of the \textit{Expectation Lower Bound (ELBO)} \citep{kingma2013}. Maximising the ELBO increases the representativeness of the learned quantised representations $z_{q}(x)$ and therefore mitigates the sampling risk with progressing model training.

\begin{figure*}[ht!]
    \center
    \includegraphics[height=5.5cm]{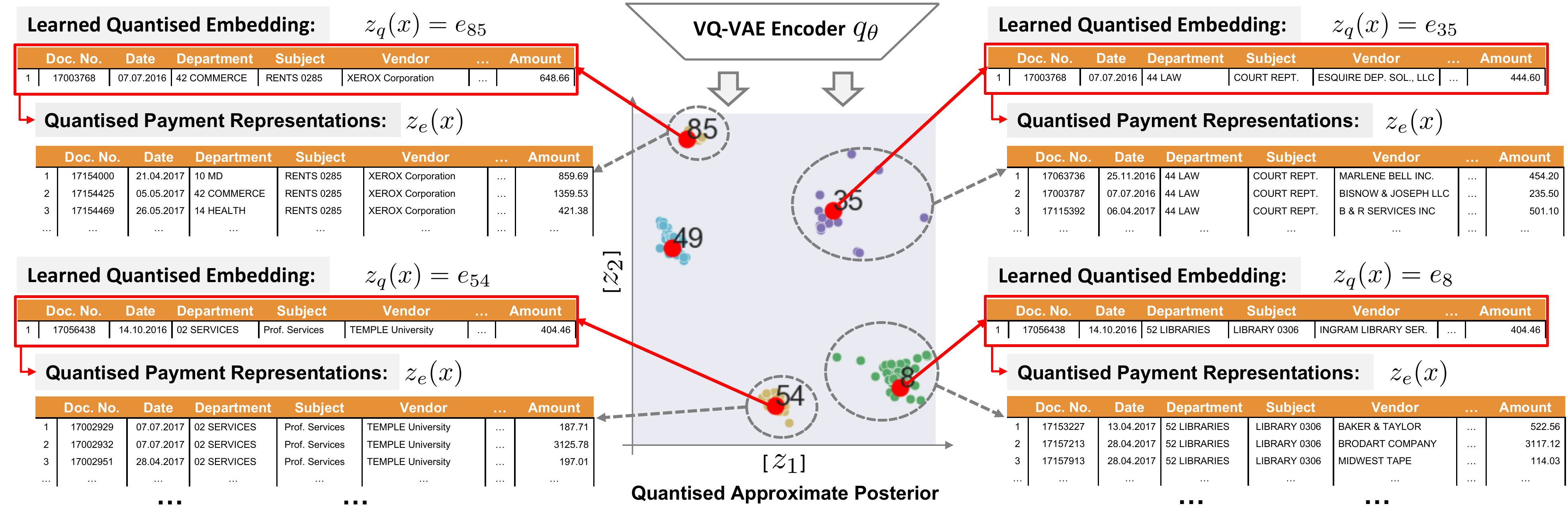}
    \caption{Exemplary VQ-VAE vector quantization of city payments and corresponding audit samples represented by the models learned embeddings $e_{k}$, for $k = \arg \min_{j} ||z_{e}(x)-e_{j}||_{2}$. For each entry $x^{i}$ VQ-VAE infers a low-dimensional representation $z_{e}$ in the latent space $Z$. The distinct representations $z_{e}$ are quantised $z_{q}$ by the embeddings $e_{k}$. As a result, the quantisations $z_{q}$ constitute a set of representative audit samples of the original entry population $X$.}
    \label{fig:manifold}
\end{figure*}

% Throughout, the learning process the representations a quantised $z_{q}$ to allow for a human interpretable audit sampling in Z.

\section{Experimental Setup}
\label{sec:experiments}

In this section, we describe the experimental setup and model training. Due to the high confidentiality of journal entry data, we evaluate the proposed methodology based on two public available real-world datasets to allow for reproducibility of our results. 

\subsection{Datasets and Data Preparation}
\label{subsec:datasets}

% In general, SAP ERP systems record journal entries and their corresponding attributes predominantly in two database tables: (1) the table "Accounting Document Headers" (technically: "BKPF") contains the meta-information of a journal entry, such as document id, type, date, time, or currency, while (2) the table  "Accounting Document Segments" (technically: "BSEG") contains the entry details, such as posting key, general ledger account, debit-credit information, or posting amount. 

To evaluate the audit sampling capability of the VQ-VAE architecture we use two publicly available datasets of real-world financial payment data that exhibit high similarity to real-world accounting data. The datasets are referred to as \textit{dataset A} and \textit{dataset B} in the following. Dataset A corresponds to the City of Philadelphia payment data of the fiscal year 2017 \footnote{The dataset is publicly available via: https://www.phila.gov/2019-03-29-philadelphias-initial-release-of-city-payments-data/.}. It represents the city's nearly \$4.2 billion in payments obtained from almost 60 city offices, departments, boards and committees. Dataset B corresponds to vendor payments of the City of Chicago ranging from 1996 to 2020 \footnote{The dataset is publicly available via: https://data.cityofchicago.org/Administration-Finance/Payments/s4vu-giwb/.}. The data is collected from the city's 'Vendor, Contract and Payment Search' and encompasses the procurement of goods and services. The majority of attributes recorded in both datasets (similar to real-world ERP data) correspond to categorical (discrete) variables, e.g. posting date, department, vendor name, document type. We pre-process the original payment line-item attributes to (i) remove of semantically redundant attributes and (ii) obtain a binary ('one-hot' encoded) representation of each payment. The following descriptive statistics summarise both datasets upon successful data pre-processing:

\begin{itemize}
\item \textbf{Dataset A:} The 'City of Philadelphia' payments encompass a total of $n=238,894$ payments comprised of $10$ categorical and one numerical attribute. The encoding resulted in a total of $8,565$ one-hot encoded dimensions for each of the city's vendor payment record $x^{i} \in \mathcal{R}^{8,565}$.

\item \textbf{Dataset B:} The 'City of Chicago' payments encompass a total of $n=72,814$ payments comprised of $7$ categorical and one numerical attribute. The encoding resulted in a total of $2,354$ one-hot encoded dimensions for each of the city's vendor payment record $x^{i} \in \mathcal{R}^{2,354}$.
\end{itemize}

\subsection{VQ-VAE Training}
\label{subsec:training}

Our architectural setup follows the VQ-VAE architecture \cite{vanDenOord2017} as shown in Fig. \ref{fig:architecture}, comprised of an encoder- and decoder-network as well as an additional embedding layer that are trained in parallel. The encoder network $q_{\theta}$ uses Leaky Rectified Linear Unit (LReLU) activation functions \cite{xu2015} except in the 'bottleneck' layer where no non-linearity is applied. The decoder network $p_{\theta}$ use LReLUs in all layers except the output layers where sigmoid activations are used. Table \ref{tab:architecture} depicts the architectural details of the networks which are implemented using PyTorch \cite{paszke2019}.

 \begin{table}[ht!]
  \caption{Number of neurons per layer $\ell$ of the encoder $q_{\theta}$ and decoder $p_{\phi}$ networks that comprise the VQ-VAE architecture \cite{vanDenOord2017} used in our experiments.} 
  \fontsize{8}{6}\selectfont
  \centering
  \begin{tabular}{l c | c c c c c c c c c}
    \toprule
        \multicolumn{1}{l}{Net}
        & \multicolumn{1}{c}{Dataset}
        & \multicolumn{1}{c}{$\ell$ = 1}
        & \multicolumn{1}{c}{2}
        & \multicolumn{1}{c}{3}
        & \multicolumn{1}{c}{4}
        & \multicolumn{1}{c}{...}
        & \multicolumn{1}{c}{10}
        & \multicolumn{1}{c}{11}
        \\
    \midrule
    $q_{\theta}(z|x)$ & A & 5,096 & 2,048 & 1,024 & 512 & ... & 4 & 2 \\
    $p_{\phi}(\hat{x}|z)$ & A & 2 & 4 & 8 & 16 & ... & 2,048 & 5,096 \\
    \midrule
    $q_{\theta}(z|x)$ & B & 2048 & 1024 & 512 & 256 & ... & 4 & 2 \\
    $p_{\phi}(\hat{x}|z)$ & B & 2 & 4 & 8 & 16 & ... & 1024 & 2048 \\
    \bottomrule \\
  \end{tabular}
    \label{tab:architecture}
 \end{table} 
 
 \vspace*{-5mm}

In accordance with \cite{xu2015}, we set the scaling factor of the LReLUs to $\alpha = 0.4$. We initialize the parameters of the encoder decoder networks as described in \cite{glorot2010}. The embeddings $e$ are initialized by sampling from a uniform prior distribution $e_{j} \sim \mathcal{U}(-1, 1)$. To allow for interpretation and visual inspection by human auditors we sample each discrete latent embedding vector $e_{j} \in \mathcal{R}^{2}$. We evaluate distinct codebook sizes of $K \in \{2^{4}, 2^{5}, 2^{6}, 2^{7}\}$ embeddings to learn different degrees of payment quantization. The models are trained with batch wise SGD for a max. of 4,000 training epochs, a mini-batch size of $m=128$ journal entries, and early stopping once the loss converges. We use Adam optimization \cite{kingma2014} with $\beta_{1}=0.9$ and $\beta_{2}=0.999$ in the optimization of the network parameters. To determine the $z_q(x)$ and $z_e(x)$ reconstruction losses (first and fourth term of Eq. \ref{equ:trainloss}) we use a Mean-Squared-Error (MSE) loss, as defined by: 

\begin{equation}
    \mathcal{L}_{MSE}(x) = ||x - p_{\phi}(q_{\theta}(x))||^{2}_{2},
    \label{equ:reconstruction_loss}
\end{equation}

\noindent where $x$ denotes an encoded journal entry, $q_{\theta}$ the encoder, and $p_{\phi}$ the decoder network. Kaiser et al. in \cite{kaiser2018} observed that training the embeddings, as done on the second term of Eq. \ref{equ:trainloss}, can be stabilized. This is achieved by maintaining an Exponential Moving Average (EMA) over (1) the embedding vectors $e_{j}$ and (2) the count $\pi_{j}$ of nearest embeddings $z_{q}(x)$ mapped onto the individual embedding vectors. Thereby, the count per embedding is defined as $\pi_{j} = \sum_{i=1}^{N} \mathds{1} [z_{q}(x^{i}) = e_{j}]$ where $\mathds{1}$ denotes the indicator function. The EMA count $c_{j}$ of each embedding is then updated per mini-batch $m$, as defined by \cite{roy2018}: 

\begin{equation}
    c_{j}^{m+1} = \eta \; c_{j}^{m} + (1 - \eta) \; \pi_{j},
    \label{equ:c_update}
\end{equation}

\noindent with updating each embedding $e_{j}$ respectively, as follows:

\begin{equation}
    e_{j}^{m+1} = \eta \; e_{j}^{m} + (1 - \eta) \sum_{i=1}^{N} \frac{\pi_{j} \; z_{e}(x^{i})}{c_{j}^{m}},
    \label{equ:e_update}
\end{equation}

\noindent where $\eta$ denotes the EMA decay parameter. We used this enhancement in all our experiments and set $\eta=0.95$. Throughout the training, we are also interested in the average number of bits used by a particular model to quantise the latent factors of variation. This is measured by codebook perplexity of the each model, as defined by:

\begin{equation}
    \mathcal{P}_{erp}(\pi) = 2^{- \sum_{j=1}^{K} p(\pi_{j}) \log_{2} p(\pi_{j})},
    \label{equ:perplexity_loss}
\end{equation}

\noindent where $p(\pi_{j})=\frac{\pi_{j}}{N}$ denotes the likelihood of a quantised representation $z_q(x)$ of being assigned to a particular embedding $e_j$. Furthermore, we determine the average purity of all payments $\omega_{j}$ quantised by a particular embedding $e_{j}$, as defined by: 

\begin{equation}
    \mathcal{P}_{urity}(\omega, \pi) = \frac{1}{K} \sum_{j=1}^{K} \frac{|\omega_{j}|}{\pi_{j}},
    \label{equ:purity_loss}
\end{equation}

\noindent where $\omega_{j} = \{x^{i} \in X | z_{q}(x^{i}) = e_{j} \}$. Figure \ref{fig:manifold} depicts an exemplary VQ-VAE vector quantization of the city payments and corresponding audit samples represented by the learned embeddings $e$. 

\section{Experimental Results}
\label{sec:results}

In this section, we first quantitatively and qualitatively assess the latent quantization learned from real-world city payments. Second, we examine the semantic disentanglement of the payments attributes in the latent dimensions in terms of interpretability by human auditors.

\subsection{Quantitative Evaluation}
\label{subsec:quantitative}

We are interested in the degree of representativeness of the learned embeddings when training VQ-VAE models with varying codebook sizes $K$. The quantitative results obtained for both datasets are shown in Tab. \ref{tab:quantise_scores}. It can be observed that with increased $K$, the average codebook usage $\mathcal{P}_{erp}$ also increases. Hence, a more fine-grained quantisation of the latent generative factors is learned. At the same time, when increasing the codebook size from $K=2^{3}$ to $2^{7}$ embeddings, the discrepancy between the (i) the quantised embeddings reconstruction loss $\mathcal{L}_{MSE}^{z_{q}}$ and (ii) the encoder output reconstruction loss $\mathcal{L}_{MSE}^{z_{e}}$ decreases. For dataset A (dataset B) the discrepancy decreases from 0.433 (0.336) to 0.141 (0.179) on average over the five random initialisations of the models. This observation corresponds to our initial hypothesis that real-world accounting data is generated by a few latent factors that can be uncovered by unsupervised deep learning. The quantitative results indicate that the VQ-VAE provides the ability to learn embeddings that quantise the latent generative factors and therefore constitute a representative audit sample.

\begin{table}[ht!]
\caption{Reconstruction losses, codebook perplexity, and cluster purity obtained for different codebook size $K$ on both city payment datasets (variances originate from parameter initialization using five random seeds).} 
\fontsize{8}{6}\selectfont
\centering
\begin{tabular}{c c | r | r | r | r }
\toprule
    \multicolumn{1}{l}{Data}
    & \multicolumn{1}{c}{$K$}
    & \multicolumn{1}{c}{$\mathcal{L}_{MSE}^{z_{q}}$}
    & \multicolumn{1}{c}{$\mathcal{L}_{MSE}^{z_{e}}$}
    & \multicolumn{1}{c}{$\mathcal{P}_{erp}$}
    & \multicolumn{1}{c}{$\mathcal{P}_{urity}$}
    \\
\midrule
A &$ 2^{3}$ &  0.748 $\pm$ 0.07 & 0.315 $\pm$ 0.07 & 6.896 $\pm$ 0.32 & 0.832 $\pm$ 0.03\\
A & $2^{4}$ & 0.746 $\pm$ 0.09 & 0.314 $\pm$ 0.02 & 13.904 $\pm$ 0.29 & 0.857 $\pm$ 0.02\\
A & $2^{5}$ & 0.614 $\pm$ 0.07 & 0.305 $\pm$ 0.03 & 23.785 $\pm$ 0.95 & 0.864 $\pm$ 0.01\\
A & $2^{6}$ & 0.553 $\pm$ 0.03 & 0.290 $\pm$ 0.05 & 40.001 $\pm$ 0.31 & 0.877 $\pm$ 0.01\\
A & $2^{7}$ & 0.436 $\pm$ 0.02 & 0.295 $\pm$ 0.03 & 54.892 $\pm$ 0.67 & 0.915 $\pm$ 0.02\\
\midrule
B & $2^{3}$ & 1.822 $\pm$ 0.05 & 1.486 $\pm$ 0.09 & 5.599 $\pm$ 0.61 & 0.543 $\pm$ 0.04\\
B & $2^{4}$ & 1.674 $\pm$ 0.04 & 1.474 $\pm$ 0.04 & 8.497 $\pm$ 0.54 & 0.575 $\pm$ 0.01\\
B & $2^{5}$ & 1.647 $\pm$ 0.07 & 1.391 $\pm$ 0.11 & 13.823 $\pm$ 0.46 & 0.593 $\pm$ 0.03\\
B & $2^{6}$ & 1.577 $\pm$ 0.08 & 1.317 $\pm$ 0.03 & 19.706 $\pm$ 1.25 & 0.634 $\pm$ 0.01\\
B & $2^{7}$ & 1.532 $\pm$ 0.05 & 1.353 $\pm$ 0.11 & 27.376 $\pm$ 2.51 & 0.727 $\pm$ 0.02\\
\bottomrule \\
\end{tabular}
\label{tab:quantise_scores}
\end{table}

\vspace*{-5mm}

\subsection{Qualitative Evaluation}
\label{subsec:qualitative}

\begin{figure*}[ht!]
    \begin{center}
        \includegraphics[height=7.0cm, width=0.3\textwidth]{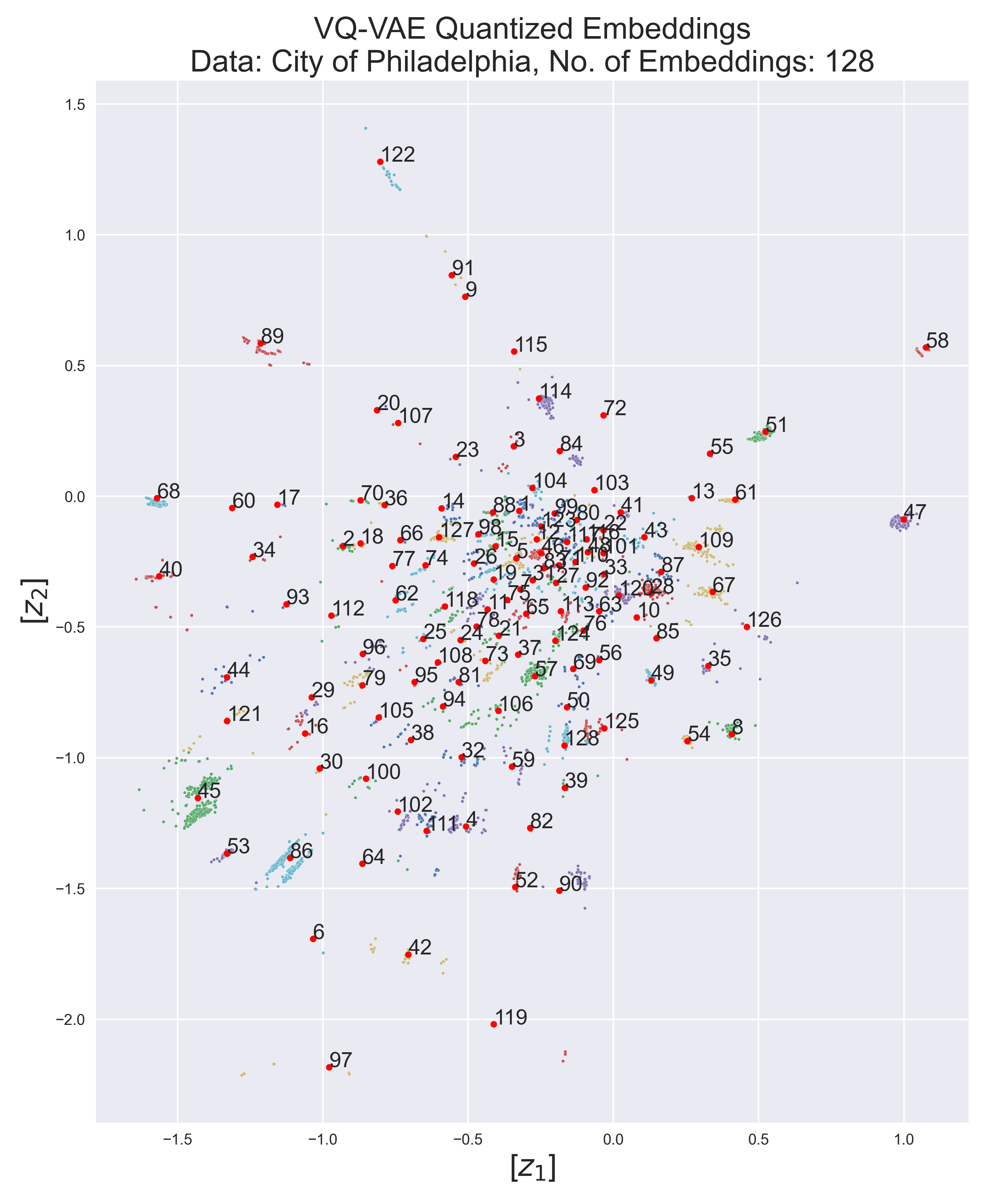}
        \includegraphics[height=7.0cm, width=0.6\textwidth]{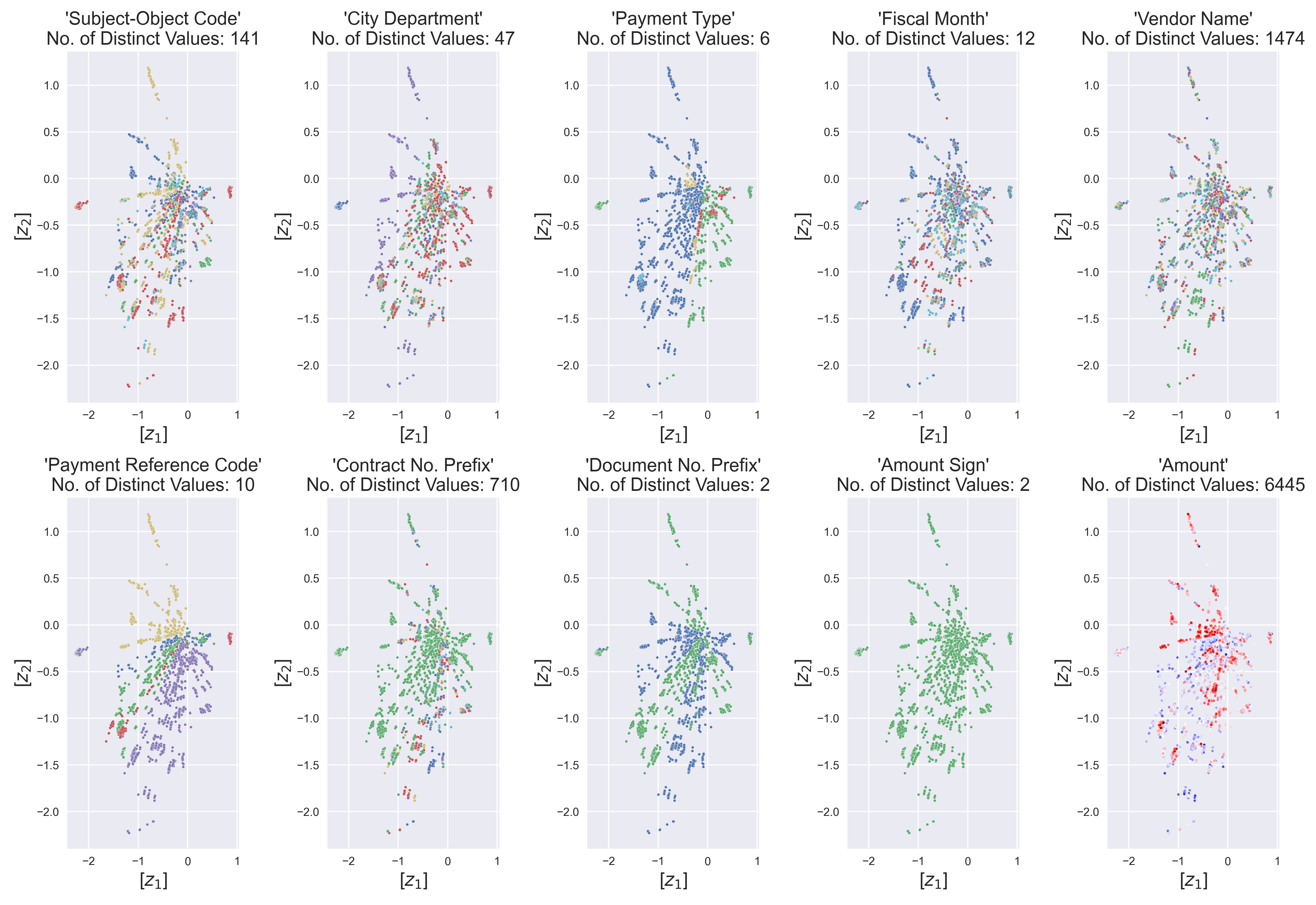}
    \end{center}
    \caption{Exemplary quantised latent representations $z_{e}$ (colored by quantisation) and embeddings $e$ (numbered red circles) in $\mathcal{R}^{2}$ learned by the VQ-VAE of the 238,894 'City of Philadelphia' vendor payments (dataset A) using a codebook size of $K=2^{7}$ (128) embeddings (left). Learned disentanglement of the distinct payment attributes $x_{j}$ (e.g., 'subject-object code', 'payment type', 'fiscal month') by the learned representations $z_{e}$ (colored by attribute value) in the latent dimensions $z_{i}$ (middle-right).}
    \label{fig:distribution_dataset_a}
\end{figure*}

\begin{figure*}[ht!]
    \begin{center}
        \includegraphics[height=7.0cm, width=0.3\textwidth]{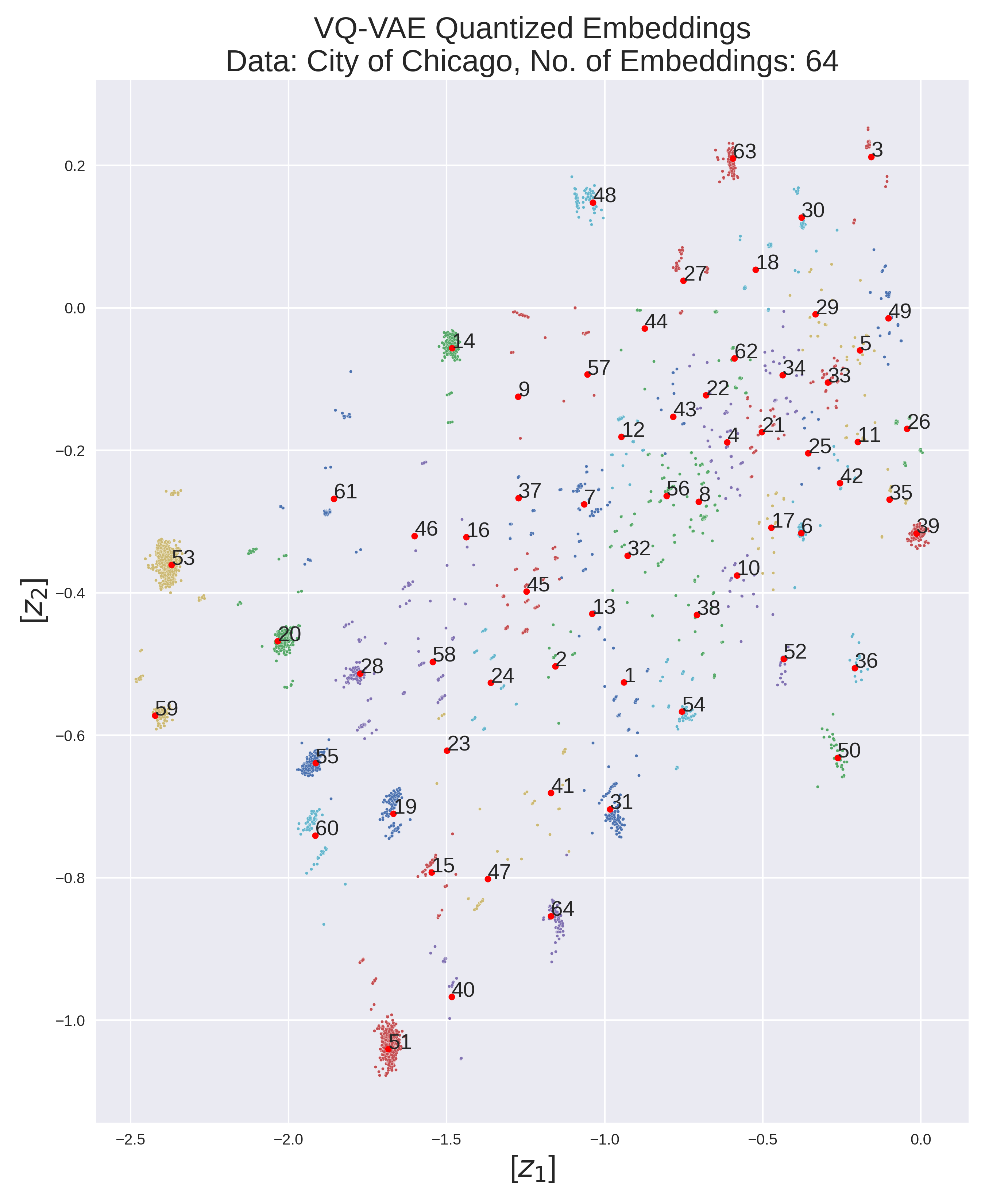}
        \includegraphics[height=7.0cm, width=0.6\textwidth]{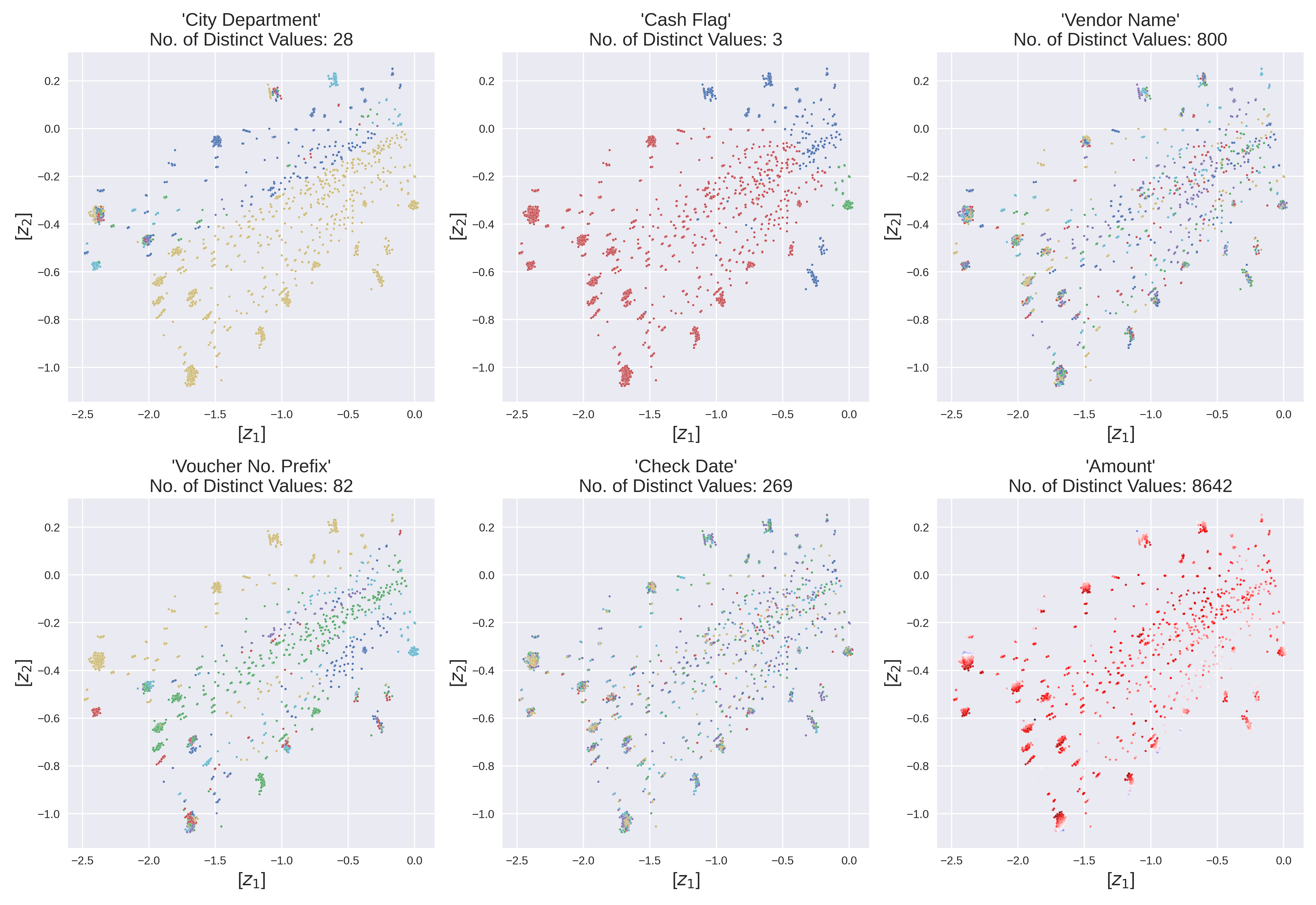}
    \end{center}
    \caption{Exemplary quantised latent representations $z_{e}$ (colored by quantisation) and embeddings $e$ (numbered red circles) in $\mathcal{R}^{2}$ learned by the VQ-VAE of the 72,814 'City of Chicago' vendor payments (dataset B) using a codebook size of $K=2^{6}$ (64) embeddings (left). Learned disentanglement of the distinct payment attributes $x_{j}$ (e.g., 'city department', 'vendor name', 'amount') by the learned representations $z_{e}$ (colored by attribute value) in the latent dimensions $z_{i}$ (middle-right).}
    \label{fig:distribution_dataset_b}
\end{figure*}

We are also interested in the semantics of the latent generative factors represented by the learned embeddings. Figures \ref{fig:distribution_dataset_a} and \ref{fig:distribution_dataset_b} show an exemplary quantisation learned by two VQ-VAE models trained with codebook sizes $K=2^{7}$ and $K=2^{6}$. For each learned embedding, we conduct a qualitative review of the corresponding quantised city payments. The review of the learned embeddings and corresponding quantised payment result in the following exemplary observations for the models with $K=2^{7}$:

% As defined in Eq. \ref{equ:lookup} each city payment is quantised by a single embedding.

\begin{itemize}

    \item \textbf{Dataset A:} Among others, the learned embeddings quantise the payments according to (1) fleet management $e_{50}$ ('auto parts'), (2) legal services $e_{52}$ ('appointed attorneys'), (3) office materials and supplies $e_{51}$ ('Staples Business Advantage'), (4) professional services $e_{121}$ ('consultancy services'), as well as (5) material and supply $e_{127}$ ('fuel and gasoline') payments;      

    \item \textbf{Dataset B:} Among others, the learned embeddings quantise the payments according to (1) transportation services $e_{2}$ ('public transport maintanance'), (2) family assistance services $e_{54}$ ('homeless financial support'), (3) aviation maintenance $e_{17}$ ('cleaning and fuel'), (4) IT services $e_{11}$ ('software'), (5) water management $e_{48}$ ('pipe supply'), as well as (5) library services $e_{60}$ ('library facilities') payments. 

\end{itemize}

\noindent The qualitative results indicate that the embeddings learned by the VQ-VAE semantically quantises the payments generative factors. The representativeness of the embeddings is also reflected by the high attribute $\mathcal{P}_{urity}$ obtainable with increased codebook size. 

\begin{table}[ht!]
\caption{Disentanglement metrics and scores obtained for both city payment datasets using different codebook sizes $K$ (variances originate from parameter initialization using five random seeds).} 
\fontsize{8}{6}\selectfont
\centering
\begin{tabular}{c c | c | c | c | c }
\toprule
    \multicolumn{1}{l}{Data}
    & \multicolumn{1}{c}{$K$}
    & \multicolumn{1}{c}{$\beta$-VAE \cite{higgins2017}}
    & \multicolumn{1}{c}{$Fac$-VAE \cite{kim2018}}
    & \multicolumn{1}{c}{MIG \cite{chen2018}}
    & \multicolumn{1}{c}{DCI \cite{eastwood2018}}
    \\
\midrule
A & $2^{3}$ & 0.153 $\pm$ 0.03 & 0.153 $\pm$ 0.02 & 0.098 $\pm$ 0.05 & 0.011 $\pm$ 0.01\\
A & $2^{4}$ & 0.155 $\pm$ 0.02 & 0.163 $\pm$ 0.03 & 0.090 $\pm$ 0.02 & 0.016 $\pm$ 0.01\\
A & $2^{5}$ & 0.192 $\pm$ 0.02 & 0.170 $\pm$ 0.01 & 0.078 $\pm$ 0.02 & 0.014 $\pm$ 0.01\\
A & $2^{6}$ & 0.216 $\pm$ 0.04 & 0.171 $\pm$ 0.02 & 0.066 $\pm$ 0.02 & 0.018 $\pm$ 0.01\\
A & $2^{7}$ & 0.242 $\pm$ 0.06 & 0.173 $\pm$ 0.01 & 0.063 $\pm$ 0.03 & 0.019 $\pm$ 0.01\\
\midrule
B & $2^{3}$ & 0.110 $\pm$ 0.01 & 0.114 $\pm$ 0.01 & 0.084 $\pm$ 0.01 & 0.042 $\pm$ 0.01\\
B & $2^{4}$ & 0.117 $\pm$ 0.01 & 0.116 $\pm$ 0.01 & 0.056 $\pm$ 0.02 & 0.045 $\pm$ 0.01\\
B & $2^{5}$ & 0.120 $\pm$ 0.01 & 0.111 $\pm$ 0.01 & 0.049 $\pm$ 0.02 & 0.046 $\pm$ 0.01\\
B & $2^{6}$ & 0.122 $\pm$ 0.01 & 0.130 $\pm$ 0.01 & 0.057 $\pm$ 0.01 & 0.048 $\pm$ 0.01\\
B & $2^{7}$ & 0.130 $\pm$ 0.02 & 0.131 $\pm$ 0.01 & 0.048 $\pm$ 0.02 & 0.047 $\pm$ 0.01\\
\bottomrule \\
\end{tabular}
\label{tab:disentanglement_scores}
\end{table}

\vspace*{-5mm}

\subsection{Disentanglement Evaluation}
\label{subsec:disentanglement}

Finally, we are interested to which extend the individual journal entry attribute characteristics are disentangled in the distinct latent dimensions $z_{i}$. A high disentanglement of the journal entry attributes, increases the interpretability of the learned quantised embeddings by a human auditor. While there is no formally accepted notion of disentanglement yet, the intuition is that a disentangled representation separates the different informative factors of variation of a dataset \citep{bengio2013a}. We evaluate four disentanglement metrics, commonly used in unsupervised representation learning \citep{locatello2019}:

\begin{itemize}

    \item The \textit{$\beta$-VAE} metric proposed in \citep{higgins2017} measures the disentanglement as the accuracy of a linear classifier that predicts the index of the fixed journal entry attribute $x_{j}$. We sample batches of 16 representations, trained the classifier on 1,000 batches, and evaluate on 500 batches.  

    \item The \textit{$Factor$-VAE} proposed in \citep{kim2018} measures the disentanglement as the accuracy of a majority vote classifier to predict the index of the fixed journal entry attribute $x_{j}$. We sample batches of 16 representations, trained the classifier on 1,000 batches, and evaluate on 500 batches.  

    \item The \textit{Mutual Information Gap (MIG)} as proposed in \citep{chen2018} measures for each journal entry attribute $x_{j}$ the two dimensions in $z_{e}(x)$ that have the highest mutual information with $x_{j}$. We bin each dimension in $Z$ into 20 bins and calculated the average mutual information over 1,000 samples.

    \item The \textit{Disentanglement Metric (DCI)} proposed in \cite{eastwood2018} computes the entropy of the distribution obtained by normalizing the importance of each dimension in $z_{e}(x)$ for predicting the value of an attribute $x_{j}$. We use a decision tree, trained the classifier on 1,000 batches, and evaluate on 500 batches. 

\end{itemize}

\noindent The disentanglement scores obtained for the distinct metrics are presented in table \ref{tab:disentanglement_scores}. It can be observed that increasing the codebook size $K$ yield an increased disentanglement of the journal entry attributes in the latent dimension of $Z$. Figure \ref{fig:distribution_dataset_a} and \ref{fig:distribution_dataset_b} and illustrate the learned disentanglement of the distinct payment attributes, e.g. city department, payment type, fiscal month. It can be observed that the learned quantised posterior disentangles the payment attributes and allows for an explainable audit sampling.

\section{Summary}
\label{sec:summary}

In this work, we proposed a deep learning inspired approach to conduct audit sampling in the context of financial statement audits. We showed that Vector Quantised-Variational Autoencoder (VQ-VAE) neural networks could be trained to learn a quantised representation of financial payment data recorded in ERP systems. We demonstrated, based on two real-world datasets of city payments, that such a learned quantisation corresponds to the latent generative factors in both datasets. Our experimental results provide initial evidence that VQ-VAE's can be utilised in the context of representative audit sampling and therefore offer the ability to reduce sampling risks. Furthermore, the learned representation allows for human interpretable discrete sampling. We hope that this technique will enhance the toolbox of auditors in the near future to sample journal entries from large-scale financial accounting data. Given the tremendous amount of journal entries recorded by organisations, deep-learning-based sampling techniques can provide a starting point for a variety of further downstream audit tasks.

% section acknowledgements
\section*{Acknowledgements}
We thank the members of the statistics department at Deutsche Bundesbank for their valuable review and remarks. Opinions expressed in this work are solely those of the authors and do not necessarily reflect the view of the Deutsche Bundesbank nor PricewaterhouseCoopers (PwC) International Ltd. and its network firms.

% section bibliography
\bibliographystyle{abbrv}
\bibliography{library}

\iffalse

\begin{table}[t!]
  \scriptsize
  \begin{center}
    \begin{tabular}{c|c|c|c|c|r|r|c|}
        \multicolumn{1}{c}{\textbf{ }}
        & \multicolumn{1}{c}{\textbf{ Company }}
        & \multicolumn{1}{c}{\textbf{ GL }}
        & \multicolumn{1}{c}{\textbf{ Profit }}
        & \multicolumn{1}{c}{\textbf{ Amount }}
        & \multicolumn{1}{c}{\textbf{ Amount}}
        & \multicolumn{1}{c}{\textbf{ Currency }}\\
        \multicolumn{1}{c}{\textbf{ }}
        & \multicolumn{1}{c}{\textbf{ Code }}
        & \multicolumn{1}{c}{\textbf{ Account }}
        & \multicolumn{1}{c}{\textbf{ Center }}
        & \multicolumn{1}{c}{\textbf{ Local }}
        & \multicolumn{1}{c}{\textbf{ Foreign }}
        & \multicolumn{1}{c}{\textbf{ Key }}\\
        \hline \hline
        \multicolumn{1}{c}{}&\multicolumn{1}{c}{}&\multicolumn{1}{c}{}\\
        1 & C20 & B24 & C1 & \textbf{882.08} & 0.00 & C1\\
        2 & C20 & B24 & C3 & \textbf{914.10} & 0.00 & C1\\
        3 & C20 & B24 & C3 & \textbf{914.10} & 0.00 & C1\\
        ... & ... & ... & ... & ... & ... & ...\\
        \multicolumn{1}{c}{}&\multicolumn{1}{c}{}&\multicolumn{1}{c}{}\\
    \end{tabular}
  \end{center}
\end{table}

\fi

\end{document}